\documentclass[letterpaper, 10 pt, conference]{ieeeconf}  

\IEEEoverridecommandlockouts                              

\usepackage[colorlinks,linkcolor=blue]{hyperref} 
\usepackage{mathrsfs}
\usepackage{amsfonts}
\usepackage{amsmath}
\usepackage[flushleft]{threeparttable}
\usepackage{tablefootnote}
\usepackage{multirow}
\usepackage{graphicx}
\usepackage{subfigure}
\usepackage{hanging}
\usepackage{color}
\usepackage{tikz}
\usetikzlibrary{calc,arrows,decorations.markings}
\usepackage{balance}
\usepackage{cite}
\usepackage{booktabs}
\usepackage{xpatch}
\usepackage{mathrsfs}
\usepackage{amsmath}
\usepackage{float}

\makeatletter
\patchcmd\@makecaption{\\}{.~}{}{\fail}
\makeatletter

\title{\LARGE \bf

GraspGPT: Leveraging Semantic Knowledge from \\ a Large Language Model for Task-Oriented Grasping}

\author{Chao Tang$^{1, 2}$, Dehao Huang$^{1, 2}$, Wenqi Ge$^{1, 2}$, Weiyu Liu$^{3}$, Hong Zhang$^{1, 2}$ \emph{Fellow, IEEE}
\thanks{$^{1}$Shenzhen Key Laboratory of Robotics and Computer Vision, Southern University of Science and Technology, Shenzhen, China.}%
\thanks{$^{2}$Department of Electronic and Electrical Engineering, Southern University of Science and Technology, Shenzhen, China.}%
\thanks{$^{3}$Institute for Robotics and Intelligent Machines, Georgia Institute of
Technology, Atlanta, United States.}%
}

\begin{document}

\maketitle


\begin{abstract}
Task-oriented grasping (TOG) refers to the problem of predicting grasps on an object that enable subsequent manipulation tasks. To model the complex relationships between objects, tasks, and grasps, existing methods incorporate semantic knowledge as priors into TOG pipelines. However, the existing semantic knowledge is typically constructed based on closed-world concept sets, restraining the generalization to novel concepts out of the pre-defined sets. To address this issue, we propose GraspGPT, a large language model (LLM) based TOG framework that leverages the open-end semantic knowledge from an LLM to achieve zero-shot generalization to novel concepts. We conduct experiments on Language Augmented TaskGrasp (LA-TaskGrasp) dataset and demonstrate that GraspGPT outperforms existing TOG methods on different held-out settings when generalizing to novel concepts out of the training set. The effectiveness of GraspGPT is further validated in real-robot experiments. Our code, data, appendix, and video are publicly available at \href{https://sites.google.com/view/graspgpt/}{https://sites.google.com/view/graspgpt}.
\end{abstract}

\thispagestyle{empty}
\pagestyle{empty}

\section{INTRODUCTION}
Tool manipulation is a fundamental skill for household robots. To achieve successful tool manipulation and accomplish specific goals, the robot must, in the first place, grasp the tool in a task-oriented manner, i.e., perform task-oriented grasping\cite{kokic2020learning, fang2020learning}. For instance, accurately gripping the handle of a knife to slice an apple into pieces or securely holding the tip of the blade of the knife during a handover. Considering the vast number of object classes and tasks in open-world operating environments like offices and kitchens, it is challenging to model the complex relationships between object classes, tasks, and grasps due to the diverse and dynamic nature of open-world environments. Practically, one can expect the robot to be trained on a limited set of examples and generalize the learned TOG skills to novel object classes and tasks beyond the training examples.


To achieve such a goal, recent works have proposed incorporating semantic knowledge into TOG pipelines to enable robots to adapt to various situations. Semantic knowledge provides high-level abstractions of open-world environments and captures the underlying relationships between concepts. For instance, Song et al.\cite{song2010learning} construct a semantic knowledge base (KB) with a pre-defined set of concepts and constraints with Bayesian Networks. Recently, Murali et al.\cite{murali2021same} contribute the largest and the most diverse TOG dataset, named TaskGrasp dataset, and build a knowledge graph (KG) based on the concepts collected in the dataset. Although these methods have demonstrated their generalization abilities to concepts pre-defined within the KB, they still operate under the closed-world assumption and cannot handle novel concepts out of the KB. This limitation is critical as a household robot must deal with open-end object classes and tasks.


\begin{figure}[t]
  \centering
  \vspace*{-0.1in}
  \begin{tikzpicture}[inner sep = 0pt, outer sep = 0pt]
    \node[anchor=south west] (fnC) at (0in,0in)
      {\includegraphics[height=4.1in,clip=true,trim=0in 0in 0in 0in]{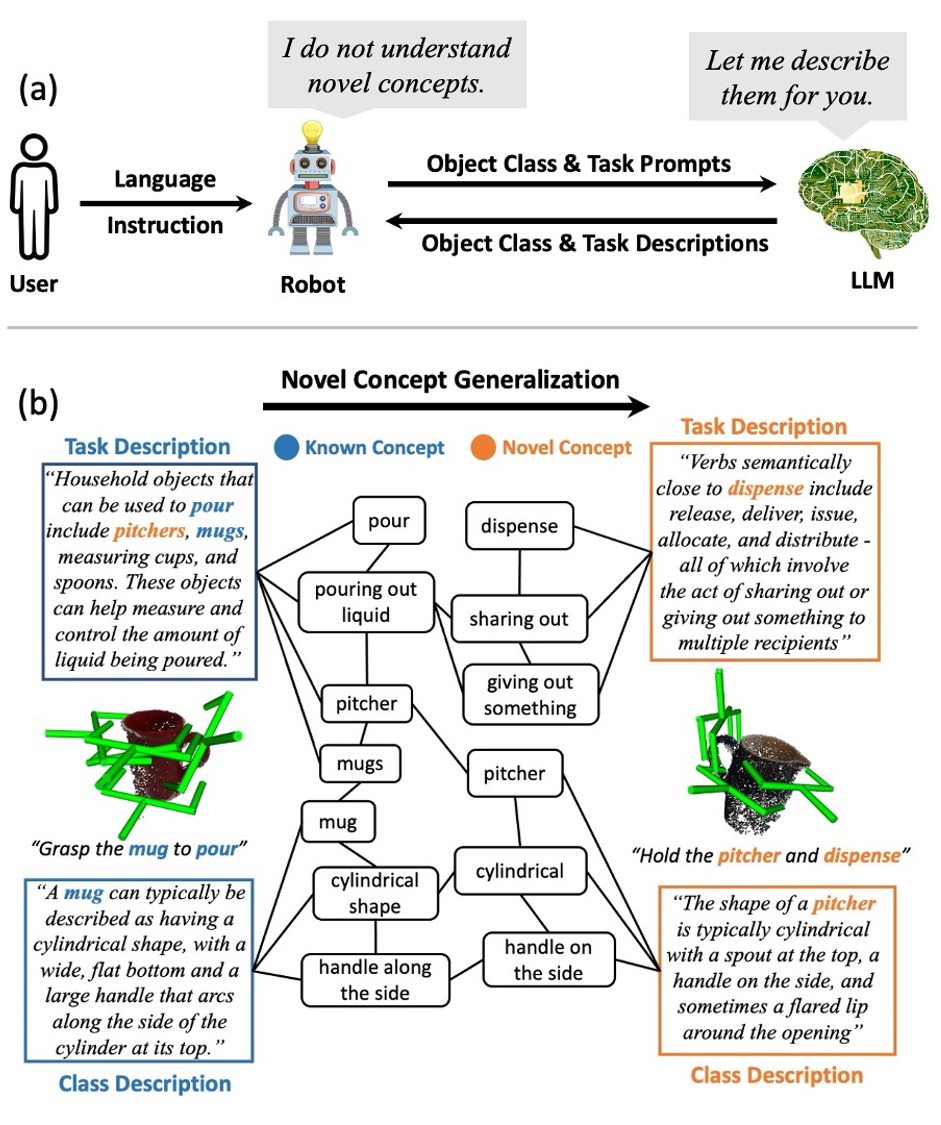}};
  \end{tikzpicture}
    \vspace*{-0.35in}
  \caption{(a) GraspGPT prompts an LLM to acquire language descriptions about the novel concept(s) in a natural language instruction given by the user. (b) Language descriptions connect the novel concept to its related concepts described during training, enabling the generalization of task-oriented grasping skills from known concepts to novel concepts.}
  \label{fig:concept}
  \vspace*{-0.3in}
\end{figure}


The recent advancements in large language models (LLMs) \cite{chowdhery2022palm, thoppilan2022lamda} have brought about significant progresses in various robot tasks\cite{ahn2022can, ren2023leveraging, huang2022language, liang2022code}. These LLMs are trained with internet-scale text corpora. Thus, robots can seamlessly extract and harness open-end semantic knowledge from LLMs to plan actions in unseen scenarios. In this letter, we follow the same spirit and introduce \textbf{GraspGPT}, an LLM-based TOG framework. GraspGPT distinguishes itself from previous TOG methods by not being constrained to a closed-world concept set. Instead, it leverages the open-end semantic knowledge about object classes and tasks from an LLM to achieve zero-shot generalization to novel concepts out of the training set. Specifically, we focus on two types of concepts: object class and task. As is shown in Figure \ref{fig:concept} (a), when presented with a novel concept in a language instruction, GraspGPT first prompts an LLM to acquire a set of natural language description paragraphs of the concept. These description paragraphs connect the novel concept to its related concepts described during training, as depicted in Figure \ref{fig:concept} (b). Subsequently, the robot can generalize the learned TOG skills from known concepts to novel concepts out of the training set. Evaluation on the contributed TOG dataset named Language Augmented TaskGrasp (LA-TaskGrasp) dataset demonstrates that GraspGPT outperforms existing TOG methods under different held-out settings. We further deploy GraspGPT on a Kinova Gen3 robotic arm to validate its effectiveness in real-world robotic applications.





In summary, our contributions are two-fold:
\begin{itemize}
    \item We propose GraspGPT, an LLM-based TOG framework that leverages the open-end semantic knowledge from an LLM to achieve zero-shot generalization to novel concepts out of the training set.
    \item We present a pipeline to automatically generate language descriptions of concepts with an LLM and contribute a language augmented TOG dataset named LA-TaskGrasp dataset.
\end{itemize}


\section{Related Work} \label{related_work}

\subsection{Task-Oriented Grasping}
The ability to perform task-oriented grasping is essential for household robots as it is the first step towards tool manipulation. Data-driven approaches have achieved success in solving TOG problems to some extent. Dang et al.\cite{dang2012semantic} and Liu et al.\cite{liu2020cage} propose novel semantic representations of grasp contexts for task-oriented grasp pose prediction. These methods learn class-task-grasp relationships purely from data without any external knowledge sources, thus achieving unsatisfying performance.


More recent works \cite{song2010learning, ardon2019learning, antanas2019semantic, murali2021same} have proposed incorporating semantic knowledge as priors into TOG pipelines. Song et al. \cite{song2010learning} construct a semantic KB with a set of tasks, object classes, actions, constraints, and reason over the KB using Bayesian Networks. Similarly, Ard{\'o}n et al. \cite{ardon2019learning} and Antanas et al. \cite{antanas2019semantic} build KGs relating pre-defined semantic attributes using probabilistic logic approaches. Despite these advancements, a significant bottleneck that hinders the generalization to a broader range of object classes and tasks is the need for large-scale TOG datasets. Motivated by this dilemma, Murali et al.\cite{murali2021same} contribute the largest and the most diverse TOG dataset, named TaskGrasp dataset, and build a KG based on the concepts collected in the dataset. More importantly, they propose the state-of-the-art TOG algorithm GCNGrasp, which builds upon the semantic knowledge encoded in the KG, to generalize to concepts within the KG. However, the major limitation of GCNGrasp is its inability to directly handle novel concepts out of the graph. This limitation is critical as a household robot must deal with open-end object classes and tasks in real-world applications. In this letter, we address this problem by leveraging the open-end semantic knowledge from an LLM to generalize learned TOG skills to novel concepts.








\subsection{LLMs in Robotics}
Recent advances in LLMs have motivated the robotics community to harness the semantic knowledge embedded in these models for a wide range of robotic applications, such as tabletop manipulation\cite{liang2022code}, navigation\cite{shah2023lm, huang2022visual}, and mobile manipulation \cite{ahn2022can, wu2023tidybot}.

Huang et al.\cite{huang2022language} first propose to decompose high-level tasks into mid-level plans with LLMs for robot decision making. To enable LLM-based robots to act properly in real-world applications, Ahn et al.\cite{ahn2022can} ground LLMs through affordance functions of pre-trained skills. Meanwhile, Huang et al.\cite{huang2022inner} extend previous work to include closed-loop feedback for both mobile and tabletop manipulation with a collection of perception models. Liang et al.\cite{liang2022code} re-purpose LLMs to directly generate policy code running on real-world robots. More recent works \cite{jiang2022vima, zeng2022socratic} combine multi-modal reasoning with LLMs for object rearrangement tasks. While aforementioned methods primarily consider LLMs for high-level task and motion planning, GraspGPT directly grounds the semantic knowledge from an LLM to grasping actions, which opens up the potential for optimizing other low-level policies (e.g., manipulation, navigation) with an LLM.

\section{Problem Formulation}\label{prob_def}

We assume access to an object class set $\mathcal{C} = \{c_i\}_{i=1}^{K_c}$ and a task set $\mathcal{T} = \{t_j\}_{j=1}^{K_t}$, where $K_c$ and $K_t$ are numbers of object classes and tasks, respectively. Based on $\mathcal{C}$ and $\mathcal{T}$, we consider the problem of learning task-oriented grasp pose prediction for a parallel-jaw gripper given the partial point cloud of an object $X_o \in \mathbb{R}^{N \times 3}$ and a natural language instruction $I$ specifying an object class $c$ and a task $t$, where $N$ is the number of points. During training, we have $c  = c_i \in \mathcal{C}$ and $t = t_j \in \mathcal{T}$. The challenge for real-world robotic applications is that $c$ or $t$ can be novel concepts (i.e., out of the training set) from open-world concept sets $\mathcal{C}_{ow}$ and $\mathcal{T}_{ow}$ during inference, where $\mathcal{C} \subset \mathcal{C}_{ow}$ and $\mathcal{T} \subset \mathcal{T}_{ow}$. To enable the generalization of TOG skills from known to novel concepts, GraspGPT incorporates open-end semantic knowledge by prompting an LLM to generate a set of object class description paragraphs $L_{c}$ for $c$ and a set of task description paragraphs $L_{t}$ for $t$.

Mathematically, we aim to estimate the posterior distribution $P(G|X_o, I, L_c, L_t)$, where $G$ represents the space of all task-oriented grasp poses. Following the convention in prior work \cite{murali2021same, mousavian20196}, the estimation process is factorized into two steps: (1) task-agnostic grasp sampling $P(G | X_o)$ and (2) task-oriented grasp evaluation $P(S|X_o, I, L_c, L_t, g)$, where $S$ is the score (probability of success) for each $g \in G$. Each grasp pose $g$ is represented by $(R, T) \in SE(3)$, where $R \in SO(3)$ represents the 3D orientation and $T \in \mathbb{R}^3$ represents the 3D translation. Since the first step is well-studied by previous works, we directly apply off-the-shelf task-agnostic grasp sampler from \cite{sundermeyer2021contact} to obtain a set of grasp pose candidates, and focus on solving the second step.



\begin{figure*}[t]
  \centering
  \begin{tikzpicture}[inner sep = 0pt, outer sep = 0pt]
    \node[anchor=south west] (fnC) at (0in,0in)
      {\includegraphics[height=2.5in,clip=true,trim=0in 0in 0in 0in]{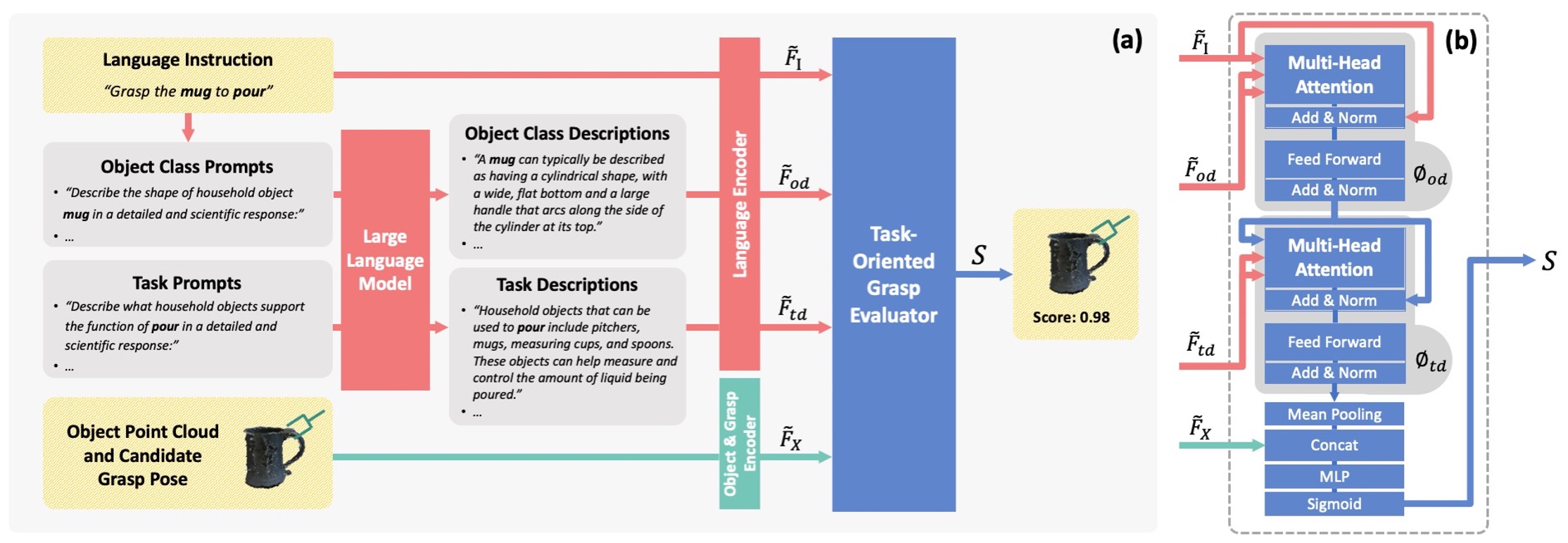}};
  \end{tikzpicture}
    \vspace*{-0.2in}
  \caption{(a) An overview of GraspGPT framework: when presented with a novel concept, such as a novel object class or task, in the natural language instruction, GraspGPT first prompts an LLM to acquire a set of language description paragraphs of the concept. Subsequently, GrasGPT evaluates the task compatibility of grasp candidates based on the multi-modal inputs from the sensors and an LLM. (b) The detailed structure of task-oriented grasp evaluator: the module is a customized transformer decoder that injects semantic knowledge from an LLM into the natural language instruction.}
  \label{fig:pipeline}
  \vspace*{-0.15in}
\end{figure*} 

\section{GraspGPT} \label{approach}

\subsection{Overview}
An overview of the proposed GraspGPT framework is presented in Figure \ref{fig:pipeline} (a). We begin by outlining the methodology for data generation in Section \ref{data+llm}. We then detail the strategy for obtaining feature representations of multi-modal inputs in Section \ref{feat}. Finally, we describe how to perform task-oriented grasp evaluation in Section \ref{tog}.

\subsection{Data Generation with an LLM}\label{data+llm}
The data generation pipeline is designed based on TaskGrasp dataset that comprises $K_c$ household object classes and $K_t$ everyday tasks. To construct the Language Augmented TaskGrasp (LA-TaskGrasp) dataset, we employ an LLM to generate language descriptions for each object class $c_i$ and task $t_j$, which will be described first. We then present the procedure for generating language instructions. \\

\begin{table*}[th]
\centering
\renewcommand\arraystretch{1.3}
\setlength\tabcolsep{10pt}
  \vspace*{-0.1in}
\begin{tabular}{ccc}
\hline
\textbf{Class} & \textbf{Property Description}                                                                                                                                                                                  & \textbf{Similarity Description}                                                                                                                                                                                                              \\ \hline \specialrule{0em}{3pt}{3pt}
\textit{Mug}   & \begin{tabular}[c]{@{}c@{}}\textit{``The mug is cylindrical in shape, with a slightly rounded} \\ \textit{base leading up to straight walls which eventually taper} \\ \textit{slightly towards the rim."}\end{tabular} & \begin{tabular}[c]{@{}c@{}}\textit{``Mugs typically have a cylindrical shape with a slightly tapered} \\ \textit{``top and a curved handle; objects with similar shapes include} \\ \textit{bottles and vases."}\end{tabular}                           \\ \specialrule{0em}{3pt}{3pt} \hline
\textbf{Task}  & \textbf{Affordance Description}                                                                                                                                                                                & \textbf{Relevance Description}                                                                                                                                                                                                               \\ \hline \specialrule{0em}{3pt}{3pt}
\textit{Sweep} & \begin{tabular}[c]{@{}c@{}}\textit{``Household objects that can be used to sweep include} \\ \textit{brooms, dustpans and mops."}\end{tabular}                                                                 & \begin{tabular}[c]{@{}c@{}}\textit{``Verbs that are semantically close to sweep include} \\ \textit{cleane, purify, and eradicate."}\end{tabular}                                                                                           \\ \specialrule{0em}{3pt}{3pt} \hline
\end{tabular}

\caption{examples of object class and task descriptions in LA-TaskGrasp dataset.}
  \vspace*{-0.3in}
\label{tab:desc_examples}
\end{table*}

\noindent \textbf{Language Description Generation} \ The key idea behind GraspGPT is to leverage the language descriptions from an LLM to establish connections between novel and known concepts. According to Rosch's theory \cite{rosch1975cognitive} of cognitive representations of semantic categories, a concept shares similar geometry, function, or effect descriptions with its related concepts. Inspired by \cite{murali2021same}, the similarity between concepts can be described by 
(1) directly prompting the LLM (e.g., \textit{``Describe what verbs are similar to cut:"}) and (2) comparing the descriptions of two concepts (e.g., \textit{``Describe the geometry of a cup/bowl:"}). In essence, the ability to relate concepts in this way is guaranteed by the fact that LLMs are trained on internet-scale data, enabling them to capture a broad range of linguistic patterns and semantic information.


To obtain $L_{c_i}$ for object class $c_i$, we design two prompt sets: (1) property prompt set, each of which asks a property of $c_i$, and (2) similarity prompt set, each of which asks classes sharing a similar property with $c_i$. Here, properties can be shape, geometry, function, etc. Similarly, to obtain $L_{t_j}$ for task $t_j$, we design: (1) affordance prompt set, each of which asks object classes that afford $t_j$, and (2) relevance prompt set, each of which asks semantically or physically relevant tasks to $t_j$. For each prompt set, we equally define $N_p$ prompts. To generate language descriptions of concepts, we recursively query the LLM to generate $N_a$ different answers per prompt. Examples of generated language descriptions are presented in Table \ref{tab:desc_examples}. We then orderly combine answers from each prompt to obtain complete description paragraphs of a concept, resulting in ${N_a}^{2N_p}$ description paragraphs. However, we have empirically observed that using only a subset of these paragraphs is sufficient for training due to the information redundancy. Each paragraph has an approximate length of 4-6 sentences. A complete list of prompts used to construct LA-TaskGrasp dataset and more examples of language descriptions can be found in the appendix. GraspGPT is not restricted to concepts defined in LA-TaskGrasp dataset, as it can incorporate open-end semantic knowledge from an LLM. It is a primary advantage over existing methods. 



\noindent \textbf{Language Instruction Generation} \ To efficiently generate language instructions during each training loop, we employ a template-based generation strategy. Following our prior work \cite{tang2023task}, we begin with $M$ templates from \cite{nguyen2020robot}, such as \textit{``Use the [obj] to [task]"}. Each template requires an object class label $c_i$ and a task label $t_j$. To further enrich the vocabulary and grammatical diversities, we perform template augmentation using an LLM (e.g., \textit{``rewrite the following sentence in a different grammatical format:"}) to generate $M^+$ additional templates, such as \textit{``hold the [obj] in your hand and [task]"} and \textit{``grip the [obj] in a [task]ing-friendly manner"}. For a complete list of $M+M^+$ templates used in LA-TaskGrasp dataset, please refer to the appendix. During each training loop, we randomly sample $c_i$ and $t_j$ from $\mathcal{C}$ and $\mathcal{T}$, and a template to generate a natural language instruction without human effort. 
    
\subsection{Multi-Modal Feature Representation}\label{feat}
To incorporate semantic knowledge about concepts in the form of language descriptions into GraspGPT framework, we transform them along with other sensory inputs into their feature representations. Therefore, two encoders are introduced: one for embedding point cloud data and the other for embedding language data. \\

\noindent \textbf{Object and Grasp Encoder} \ To model the relative spatial relationship between a 6 DoF (Degree of Freedom) grasp pose and $X_o$, we adopt a joint embedding strategy. Following Mousavian et al. \cite{mousavian20196}, we first approximate the robot gripper with six control points $X_g$ defined in the object frame and concatenate them to the object point cloud $X_o$ to form a joint point cloud. A binary feature vector is then added to the joint point cloud, indicating that each point belongs to the object or the gripper. Finally, the joint point cloud is embedded with PointNet++\cite{qi2017pointnet++} (denoted as PN++), which consists of three set abstraction layers:
\begin{align*}
    F_X = \textnormal{PN++}(\textnormal{Concat}([X_g, X_o], \textnormal{dim}=0))
\end{align*}
The resulting point cloud embedding $F_X \in \mathbb{R}^{1024}$ is later fused with language embeddings. \\



\noindent \textbf{Language Encoder} \ In order to relate known and novel concepts, GraspGPT necessitates the ability to digest a large variety of linguistic elements in language descriptions. For instance, in the case of an affordance description of the task \textit{``pour"}:
\begin{center}

``\textit{Household objects that support the function of pouring include utensils such as pitchers, cups, and ladles, as well as containers with pouring spouts, to aid in the transfer of liquid or other items from one vessel to another.}"
    
\end{center}
GraspGPT must be able to comprehend object classes (e.g., \textit{pitchers}, \textit{cups}, \textit{ladles}), entity taxonomy (e.g., \textit{utensil}, \textit{container}, \textit{vessel}, \textit{liquid}), actions (e.g., \textit{transfer}), and relations (e.g., \textit{from ... to ...}). While training a dedicated language encoder from scratch is a common choice, it would require a significant amount of training data and is meanwhile time-consuming. We, therefore, opt for a BERT\cite{devlin2018bert} pre-trained on a large corpus of text data to encode both language descriptions and instructions. The pre-trained BERT outputs word embeddings for a task description paragraph $F_{td} \in \mathbb{R}^{T_{td} \times 768}$, an object class description paragraph $F_{od} \in \mathbb{R}^{T_{od} \times 768}$, and a language instruction $F_{I} \in \mathbb{R}^{T_{I} \times 768}$, where $T_{td}$, $T_{od}$, and $T_{I}$ denote the maximum lengths (with zero-padding) for each language sequence, respectively. The language encoder is frozen during training.

\subsection{Task-Oriented Grasp Evaluation}\label{tog}
After obtaining the feature representations of all elements, we next present a multi-modal fusion module for task-oriented grasp evaluation, which can be represented below:
\begin{align*}
    S = \textnormal{TGE}(F_X, F_I, F_{td}, F_{od})
\end{align*}
where TGE is the task-oriented grasp evaluator, and $S$ is the score for the candidate grasp pose $g$. \\



\noindent \textbf{Task-Oriented Grasp Evaluator} \ TGE is implemented as a customized Transformer decoder\cite{vaswani2017attention}. It is analogous to a sequence-to-sequence model commonly used in machine translation, which converts sequences from one domain to another. In our problem, the robot is unable to comprehend novel concepts out of the training set. We utilize TGE to translate a novel concept using its description paragraphs, and connect the novel concept to its related concepts described during training.

The architecture of TGE is depicted in Figure \ref{fig:pipeline} (b). The translation process incorporates contextual information from language descriptions into their corresponding concept in $I$. Both training and inference follow the same computational procedure. Specifically, we begin by transforming word embeddings from the pre-trained language encoder to a lower dimension space and obtain $\Tilde{F}_{td} \in \mathbb{R}^{T_{td} \times 128}$, $\Tilde{F}_{od}\in \mathbb{R}^{T_{od} \times  128}$, and $\Tilde{F}_{I} \in \mathbb{R}^{T_I \times 128}$. TGE consists of two layers, one for incorporating object class knowledge $\Tilde{F}_{od}$ and the other for incorporating task knowledge $\Tilde{F}_{td}$. Each decoder layer aims to learn a function as below:
\begin{align*}
    \phi_{td}: \mathbb{R}^{T_{I} \times 128} \times \mathbb{R}^{T_{td} \times 128} \rightarrow \mathbb{R}^{T_{I} \times 128} \\
    \phi_{od}: \mathbb{R}^{T_{I} \times 128} \times \mathbb{R}^{T_{od} \times 128} \rightarrow \mathbb{R}^{T_{I} \times 128}
\end{align*}
where the outputs of $\phi_{td}$ and $\phi_{od}$ are language instruction word embeddings augmented with contextual knowledge. Two decoder layers share a similar design. The computational procedure of $\phi_{*d}$ can be represented as follows:
\begin{align*}
    \Tilde{F}_{I} &= \text{LN}(\Tilde{F}_{I} + \text{MHA}(\Tilde{F}_{I}, \Tilde{F}_{*d})) \\
    \Tilde{F}_{I} &= \text{LN}(\Tilde{F}_{I} + \text{FFN}(\Tilde{F}_{I}))
 \end{align*}
where $*d$ can be either $td$ or $od$; LN, MHA, and FFN denote layer normalization, multi-head attention, and feedforward network, respectively; MHA consists of eight cross-attention heads in our implementation. The computation of each cross-attention head can be represented as:
\begin{align*}
    A = \text{Softmax}(\frac{Q_IK_{*d}^{T}}{\sqrt{128}})V_{*d}
\end{align*}
where $A$ is the attended word embeddings. $Q_I$, $K_{*d}$, and $V_{*d}$ are transformed from $\Tilde{F}_{I}$ and $\Tilde{F}_{*d}$ as follows:
\begin{align*}
    Q_I = \textnormal{Q}_{\textnormal{proj}}(\Tilde{F}_{I}), K_{*d} = \textnormal{K}_{\textnormal{proj}}(\Tilde{F}_{*d}), V_{*d} = \textnormal{V}_{\textnormal{proj}}(\Tilde{F}_{*d}) 
\end{align*}
where $\textnormal{Q}_{\textnormal{proj}}$, $\textnormal{K}_{\textnormal{proj}}$, and $\textnormal{V}_{\textnormal{proj}}$ are projection matrices. The intuition is to reconstruct $\Tilde{F}_{I}$ by all elements in $\Tilde{F}_{*d}$ weighted by their normalized correspondence. Since cross-attention mechanism can dynamically assign weights to each input token, it learns to attend to concept tokens in $I$ while ignore irrelevant tokens. 

Finally, $\Tilde{F}_{I}$ is mean pooled to output a sentence embedding $\overline{F}_{I}\in \mathbb{R}^{128}$. It is then concatenated with the shape embedding $\Tilde{F}_{X} \in \mathbb{R}^{300}$, which is obtained by projecting $F_X$ via a fully connected layer. We compute $S$ using an MLP (Multi-Layer Perceptron) with a sigmoid activation:
\begin{align*}
    S = \textnormal{Sigmoid}(\textnormal{MLP}(\textnormal{Concat}([\Tilde{F}_{X}, \overline{F}_{I}], \textnormal{dim}=-1)))
\end{align*} 
The MLP comprises three fully connected layers with 1D batch normalization, ReLU activation, and dropout.\\

\noindent \textbf{Loss Function} \ We compute the binary cross-entropy loss between $S$ and the ground truth label $S_{gt}$:
\begin{align*}
    \mathcal{L}_{bce} = -\frac{1}{N} \sum_{i=1}^{N} S_{gt, i} \cdot \text{log}(S_{i}) + \\ (1 - S_{gt, i}) \cdot \text{log}(1-S_{i})
\end{align*}
where $N$ is the total number of samples, and $S_{gt, i}$ is set to one if the $i_{th}$ grasp pose is successful and zero otherwise.

\section{Experimental Setup} \label{exp_setup}

\subsection{Perception Experiments} \label{perception_exp_setup}
\noindent \textbf{Baselines} \  We compare GraspGPT to the following methods: (1) \textbf{Random}, which represents the method in \cite{sundermeyer2021contact} that focuses on grasp stability only and ignores task constraints (i.e., task-agnostic grasping method). (2) \textbf{Semantic Grasp Network (SGN)} \cite{liu2020cage}, which learns class-task-grasp relations without incorporating external semantic knowledge. (3) \textbf{GCNGrasp} \cite{murali2021same}, which is the state-of-the-art TOG algorithm introduced earlier and whose main limitation is its inability to generalize to novel concepts out of the graph. During inference, we connect the novel concept node to its nearest neighbor in the KG. The nearest neighbor search is based on the cosine similarity between the concepts' pre-trained word embeddings provided by ConceptNet \cite{liu2004conceptnet}. \\


\noindent \textbf{Dataset} \ GraspGPT and three baselines are evaluated on the LA-TaskGrasp dataset, which augments the TaskGrasp dataset with language data. The original TaskGrasp dataset contains 250K task-oriented grasp pose annotations for 56 tasks, 75 object classes, and 191 object instances. Each instance is a partial point cloud of a real household object \textbf{with multi-view RGB-D fusion}. TaskGrasp provides three types of held-out settings: held-out (object) class, held-out task, and held-out instance. We focus on the former two settings in this letter. For language data, LA-TaskGrasp contains 80 language description records for each object class and 40 records for each task, resulting in 6000 object class description records and 2240 task description records. We combine these descriptions to generate 750 object class description paragraphs and 560 task description paragraphs. LA-TaskGrasp dataset also includes 53 language instruction templates, resulting in 222600 possible language instruction sentences. \\


\begin{figure*}[t]
  \centering
  \begin{tikzpicture}[inner sep = 0pt, outer sep = 0pt]
    \node[anchor=south west] (fnC) at (0in,0in)
      {\includegraphics[height=2.2in,clip=true,trim=0.3in 0in 0in 0in]{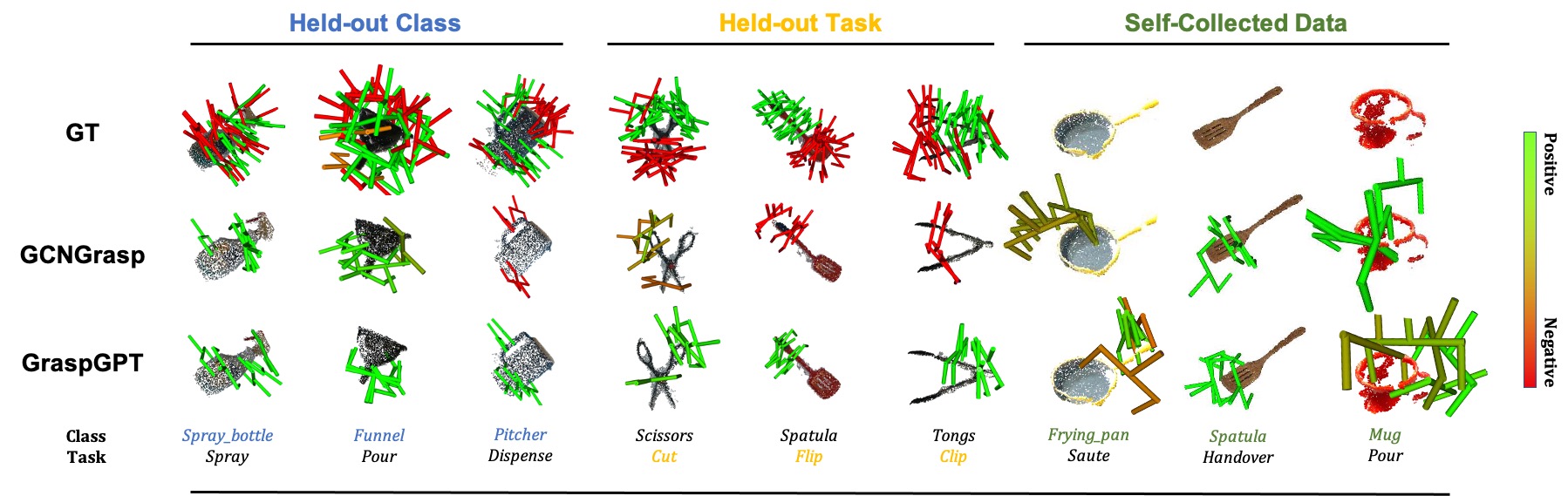}};
  \end{tikzpicture}
  \caption{Qualitative results of open-world generalization. GraspGPT and GCNGrasp are evaluated under both held-out settings. Results on self-collected test objects (no ground truth annotations) are also presented. Grasp poses are colored by their confidence scores (green is higher). Only top-5 predictions are displayed for better visualization effect.}
  \label{fig:qualitative}
\end{figure*} 

\noindent \textbf{Metrics} \  We use the same set of evaluation metrics used by GCNGrasp. Specifically, we compute the Average Precision (AP) score for each object class, task, and instance, and then compute the mean AP (mAP) averaged over all object classes, tasks, and instances (i.e., class mAP, task mAP, and instance mAP).

\subsection{Real-Robot Experiments}
The real-robot experiment platform comprises a 7 DoF Kinova Gen3 robotic arm with a parallel jaw gripper and an Intel RealSense D435 RGB-D camera with eye-in-hand calibration. For each test object, we first apply SAM\cite{kirillov2023segment} to extract the object point cloud \textbf{captured from a single view} and then apply Contact-GraspNet\cite{sundermeyer2021contact} to generate 50 grasp pose candidates. Single-view setup is used here because it is more practical for real-world robotic applications. We collect test objects from our laboratory and YCB dataset. More details on the experimental setup can be found in the appendix.

The physical grasping pipeline is divided into three stages: Perception, Planning, and Action, and the statistics of each stage is reported separately for clarity. A trial succeeds if the test object is grasped subject to the natural language instruction and lifted stably by the robot. We additionally combine GraspGPT with three pre-defined skills (pouring, handover, and scooping) in the form of motion primitive to showcase its practicality in tool manipulation. 




\subsection{Implementation Details}\label{detail}
All the experiments are conducted on a desktop PC with a single Nvidia RTX 3090 GPU. GraspGPT is optimized with an Adam optimizer\cite{kingma2014adam} with a weight decay of 0.0001. The learning rate is set to 0.0001 initially and decays subject to a customized function as in GCNGrasp. We train GraspGPT for 50 epochs with a batch size of 32. Each point cloud is downsampled to 4096 points before being fed into the model. 


For the choice of an LLM, we select the OpenAI GPT-3 model, specifically the \textit{text-davinci-003} version. GraspGPT is capable of incorporating any current LLM, such as OpenAI GPT-4 and Google Bard, or using an ensemble of LLMs. We leave this ensemble approach for our future work. For the language encoder, we choose the Google pre-trained \textit{BERT-Base} model provided by Hugging Face.

\begin{table*}[th]
\centering
\renewcommand\arraystretch{1.6}
\setlength\tabcolsep{20pt}
\begin{tabular}{ccccccc}
\toprule
\multirow{2}{*}{\textbf{Method}} & \multicolumn{3}{c}{\textbf{Held-out Class Performance (mAP)}} & \multicolumn{3}{c}{\textbf{Held-out Task Performance (mAP)}} \\ \cline{2-7} 
                                                & Instance           & Class          & Task           & Instance          & Class          & Task           \\ \hline
Random                  & 59.32              & 58.73          & 52.72          & 59.06             & 58.24          & 52.37          \\
SGN                     & 74.20              & 72.95          & 62.55          & 75.17             & 71.59          & 63.35          \\ \hline
GCNGrasp (open-world)   & 72.92              & 72.45          & 67.58          & 57.48             & 47.21          & 33.17          \\
GCNGrasp (closed-world) & 79.35              & 76.88          & 72.97          & 80.43             & 76.06          & 76.11          \\ \hline
GraspGPT (full model)   & 79.70              & 77.88          & 72.84          & 79.32             & 76.90          & 72.34          \\
GraspGPT (w/o D)        & 74.10              & 74.33          & 66.38          & 74.66             & 70.85          & 68.14          \\
GraspGPT (w/o TD)       & 80.95              & 77.74          & 73.76          & 75.00             & 71.21          & 68.38          \\
GraspGPT (w/o OD)       & 76.04              & 74.71          & 71.60          & 78.26             & 74.71          & 71.60          \\\bottomrule
\end{tabular}
\caption{quantitative results of perception experiments}
\label{tab:in-out-eval}
  \vspace*{-0.15in}
\end{table*}


\begin{table*}[t]
\centering
\renewcommand\arraystretch{1.6}
\setlength\tabcolsep{12pt}
\begin{tabular}{ccccccccc}
\toprule
\multirow{2}{*}{\textbf{Method}} & \multicolumn{3}{c}{\textbf{Held-out Class Performance}} & \multirow{2}{*}{Success} & \multicolumn{3}{c}{\textbf{Held-out Task Performance}} & \multirow{2}{*}{Success} \\ \cline{2-4} \cline{6-8}
                        & Perception          & Planning         & Action         &                          & Perception          & Planning         & Action         &                          \\ 
\hline
GraspGPT   & 91/100             & 85/100            & 77/100           & 77.00\%                       & 86/100             & 77/100             & 71/100           & 71.00\%                      \\ \bottomrule
\end{tabular}
\caption{results of task-oriented grasping experiments}
\label{tab:real_grasp}
  \vspace*{-0.15in}
\end{table*}

\begin{table*}[t]
\centering
\renewcommand\arraystretch{1.6}
\setlength\tabcolsep{8pt}
\begin{tabular}{cccccccccc}
\toprule
\multirow{2}{*}{\textbf{Method}} & \multicolumn{2}{c}{\textbf{Pouring}} & \multirow{2}{*}{Success} & \multicolumn{2}{c}{\textbf{Handover}} & \multirow{2}{*}{Success} & \multicolumn{2}{c}{\textbf{Scooping}} & \multirow{2}{*}{Success} \\ \cline{2-3} \cline{5-6} \cline{8-9}
                        & Grasping   & Manipulation   &                          & Grasping    & Manipulation   &                          & Grasping    & Manipulation   &                          \\ \hline
GraspGPT                & 15/20      & 12/20         & 60.00\%                  & 17/20       & 16/20          & 80.00\%                  & 18/20       & 13/20          & 65.00\%                  \\ \bottomrule
\end{tabular}
\caption{results of task-oriented manipulation experiments}
\label{tab:real_mani}
  \vspace*{-0.3in}
\end{table*}

\section{Results} \label{exp}

\subsection{Results of Perception Experiments}
To highlight the difference between our approach and GCNGrasp, we investigate perception experiments from two perspectives: open-world generalization and closed-world generalization. The former evaluates the generalization performance to novel concepts out of the knowledge graph of GCNGrasp. In the latter evaluation, GCNGrasp has access to all the concepts in the LA-TaskGrasp dataset and the ground truth relations between them in its pre-defined graph. Although this assumption is impractical in real-world robotic applications, we still want to explore how our approach compares to GCNGrasp even though GraspGPT does not assume access to concepts out of the training set. Since GraspGPT and the other two baselines do not rely on a pre-defined graph structure, their results for the two evaluations are the same. The quantitative results of perception experiments are reported in Table \ref{tab:in-out-eval}. 

\noindent \textbf{Open-World Generalization} \ For both held-out settings, Random achieves approximate mAPs of 50-60\%, indicating that the distribution of positive and negative samples in the dataset is even. By considering task constraints, SGN achieves consistent improvements (10\%+) over Random under two held-out settings. For GCNGrasp, we observe a significant performance difference between the held-out task setting and the held-out class setting. GCNGrasp even falls behind Random by 1.58\%, 11.03\%, and 19.20\% on three metrics in the held-out task setting. This suggests that the pre-trained word embeddings of ConceptNet are good at capturing the linguistic relations between object classes but perform poorly on relating task concepts. Therefore, GCNGrasp cannot fully exploit the power of semantic knowledge encoded in its graph. Since GraspGPT does not rely on a pre-defined KG but instead leverages the open-end semantic knowledge from an LLM, \textbf{GraspGPT outperforms all three baselines when generalizing to concepts out of the training set}. It outperforms GCNGrasp by 21.84\%, 29.69\%, and 39.17\% on held-out task setting and by 6.78\%, 5.43\%, and 5.26\% on held-out class setting. The qualitative results are presented in Fig. \ref{fig:qualitative}. \\

\noindent \textbf{Closed-World Generalization} \ Compared to open-world generalization, GCNGrasp performs consistently better on closed-world generalization since all the concepts and the ground truth relations between them have been pre-defined in its graph. GraspGPT and GCNGrasp outperform both Random and SGN due to the incorporation of semantic knowledge. For the held-out task setting, GraspGPT achieves comparable performance with GCNGrasp on instance mAP and class mAP but falls behind by 3.77\% on task mAP. For held-out class setting, GraspGPT outperforms GCNGrasp on two metrics. Overall, \textbf{GraspGPT achieves comparable performance with GCNGrasp on closed-world generalization} even though it does not assume access to all concepts and their relations as GCNGrasp does.

\subsection{Results of Real-Robot Experiments}

\noindent \textbf{Task-Oriented Grasping} \ We conduct 100 trials on each held-out setting, with ten trials per object class or task. As presented in Table \ref{tab:real_grasp}, GraspGPT achieves high success rates (86.00\% and 91.00\%) in the perception stage, even though the object point clouds are captured from a single view. The performance drop from the perception stage to the action stage (71.00\% and 77.00\%) can be attributed to three primary reasons: (1) marginal grasp candidates generated by the grasp sampler; (2) incorrect evaluation by GraspGPT; (3) motion planning failure. The qualitative results of three test objects are shown in Figure \ref{fig:qualitative} (right). \\

\begin{figure}[t]
  \centering
  \begin{tikzpicture}[inner sep = 0pt, outer sep = 0pt]
    \node[anchor=south west] (fnC) at (0in,0in)
      {\includegraphics[height=2.4in,clip=true,trim=0in 0in 0in 0in]{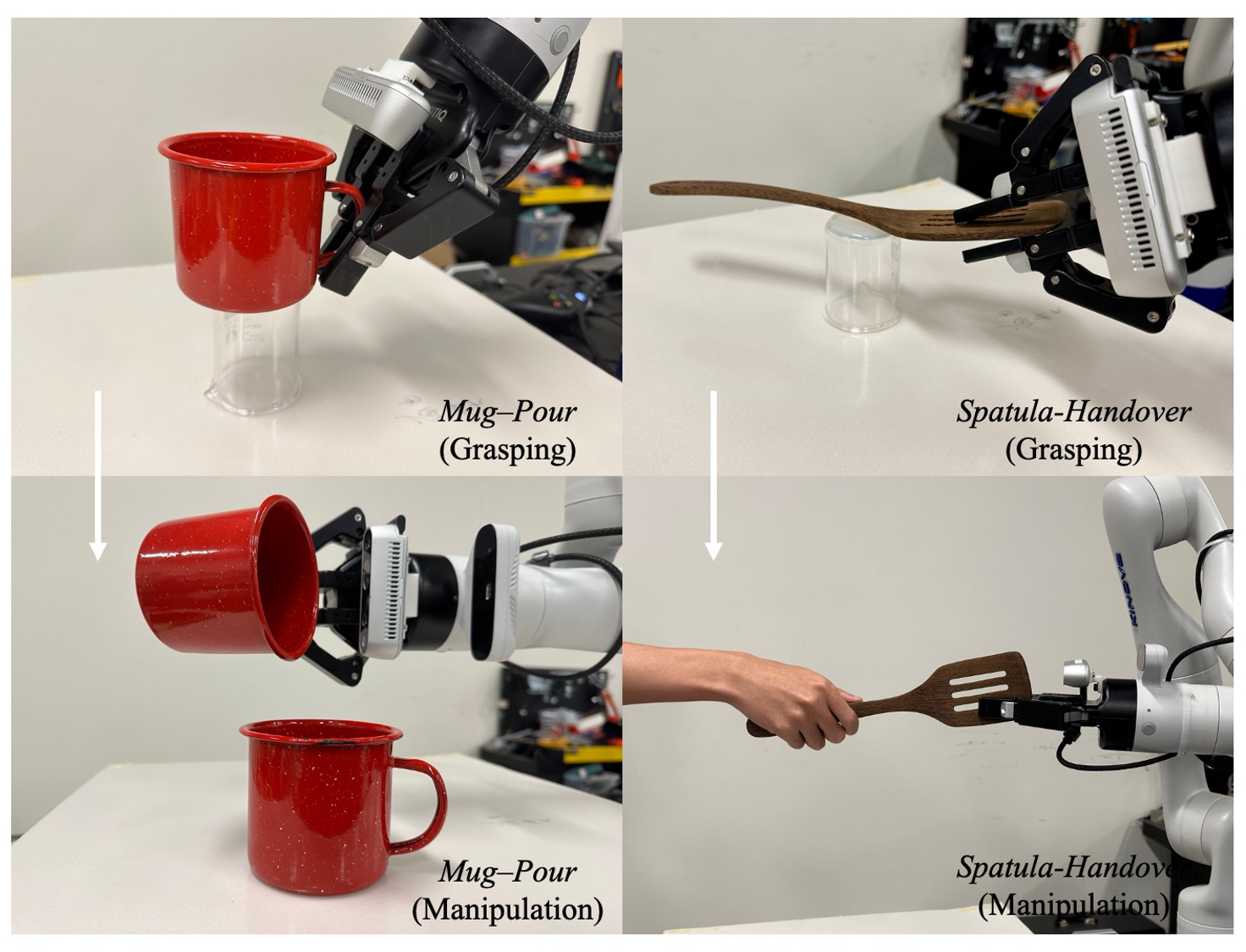}};
  \end{tikzpicture}
    \vspace*{-0.1in}
  \caption{Real-robot experiments on task-oriented grasping and manipulation: \textit{Mug-Pour} (left) and \textit{Spatula-Handover} (right).}
  \label{fig:real-robot}
  \vspace*{-0.25in}
\end{figure}

\noindent \textbf{Task-Oriented Manipulation} \ To support task-oriented manipulation (refer to Figure \ref{fig:real-robot}), we first utilize GraspGPT to generate task-oriented grasp poses for tool objects. Then, we design rule-based heuristics to determine the operating direction and effect points \cite{qin2020keto} on the target objects. As presented in Table \ref{tab:real_mani}, GraspGPT performs well in task-oriented grasping, achieving success rates of 75.00\%, 85.00\%, and 90.00\% in three tasks, respectively. However, due to its inability to adaptively model the relative pose \cite{pan2023tax} between the tool object and the target object, the success rates of task-oriented manipulation decrease, especially for pouring and scooping.


\subsection{Ablation Study}\label{augment_exp}
To gain further insights into the effectiveness of each component of GraspGPT, we perform two sets of ablation studies, aiming to answer two questions:
\begin{itemize}
    \item Does the incorporation of semantic knowledge from an LLM help to better generalize to novel concepts out of the training set?
    \item How does the selection of a pre-trained language encoder affect the overall performance of GraspGPT?    
\end{itemize}


\noindent \textbf{Ablation on Semantic Knowledge} \ We compare GraspGPT to three ablations: (1) no semantic knowledge (i.e., w/o D); (2) object class description only (i.e., w/o TD); (3) task description only (i.e., w/o OD). The results for two held-out settings are reported in Table \ref{tab:in-out-eval}. For the held-out task setting, the full model outperforms all three ablations. Specifically, the comparison between w/o D - w/o TD and w/o D - w/o OD demonstrates that the incorporation of task knowledge is more important for novel task generalization. For the held-out class setting, we observe that object class knowledge is more important for generalizing to novel classes out of the training set. Using object class knowledge only (w/o TD) even slightly outperforms the full model. We argue that object class descriptions have already provided sufficient knowledge for novel object class generalization. Arbitrarily incorporating extra task knowledge may lead to adversarial/conflicting effects in some cases. In our current implementation, we do not preprocess the language data from the LLM. Future work will be done on knowledge filtering and selection. Overall, the result verifies the hypothesis that \textbf{the incorporation of semantic knowledge helps achieve better generalization to novel concepts.} \\



\noindent \textbf{Ablation on Language Encoder} \ To validate the design choice of using a large pre-trained language encoder, we equip GraspGPT with pre-trained BERTs of three sizes and compare their resulting mAPs. Since the conclusions for the two held-out settings are similar, we only report the result of the held-out task setting for simplicity. The result is presented in Table \ref{tab:ablation_le}, where $L$ denotes the number of transformer layers. It is clear that \textbf{\textit{BERT-Base} outperforms two smaller models}, but the gaps are insignificant. We argue that the three models are equally pre-trained on a large corpus of text data, so they achieve a similar level of knowledge understanding capability despite their differences in model complexity. 

\begin{table}[t]
\centering
\renewcommand\arraystretch{1.6}
\setlength\tabcolsep{12pt}
\begin{tabular}{cccc}
\toprule
\multirow{2}{*}{\textbf{Model}} & \multicolumn{3}{c}{\textbf{Held-out Task Performance (mAP)}} \\ \cline{2-4} 
                         & Instance      & Class      & Task     \\ \hline
BERT-Small ($L$=4)               & 78.06             & 74.48           & 71.90         \\
BERT-Medium ($L$=8)              & 78.47             & 75.93           & 72.02         \\ \hline
BERT-Base ($L$=12)                & \textbf{79.32}          & \textbf{76.06}        & \textbf{72.34} \\ \bottomrule
\end{tabular}
\caption{ablation on pre-trained language encoder}
\label{tab:ablation_le}
  \vspace*{-0.3in}
\end{table}

\section{Conclusion} \label{conclusion}
In this letter, we propose GraspGPT, an LLM-based TOG framework that leverages the open-end semantic knowledge from an LLM to achieve zero-shot generalization to novel concepts out of the training set. Compared to existing methods, GraspGPT does not rely on any pre-defined concept set or knowledge base. Evaluation on the LA-TaskGrasp dataset demonstrates the superiority of GraspGPT over existing methods on novel concept generalization. The effectiveness of GraspGPT is further validated in performing task-oriented grasping and manipulation in real-world applications.


\section{Acknowledgments}
This work was supported in part by the Shenzhen Key Laboratory of Robotics and Computer Vision (ZDSYS20220330160557001).







\bibliographystyle{IEEEtran}
\balance
\bibliography{main}

\newpage

\onecolumn

\begin{appendices}

\section{}

\subsection{Language Augmented TaskGrasp (LA-TaskGrasp) Dataset}

This section presents the natural language prompts utilized to generate object class and task descriptions within the LA-TaskGrasp dataset. Additionally, we provide detailed examples of object class descriptions (\textit{``Mug"}, \textit{``Spoon"}, and \textit{``Hammer"}) and task descriptions (\textit{``Sweep"}, \textit{``Screw"}, and \textit{``Slice"}). Following that, we present a list of 53 language instruction templates, consisting of 11 templates derived from previous work and 42 templates generated through LLM data augmentation. Lastly, we offer 18 LA-TaskGrasp dataset examples, each of which includes 6 DoF task-oriented grasp poses, a language instruction, an object class description paragraph, and a task description paragraph.\\

\subsubsection{Object Class Description Prompts} 

\begin{itemize}
    \setlength{\itemsep}{5pt}
    \item ``\textit{Describe the shape/geometry of household object [obj] in a detailed and scientific response:}"
    \item ``\textit{Describe the common use/function of household object [obj] in a detailed and scientific response:}"
    \item ``\textit{Describe what household objects have similar shapes/geometries to [obj] in a detailed and scientific response:}"
    \item ``\textit{Describe what household objects have similar uses/functions to [obj] in a detailed and scientific response:}" \\
\end{itemize} 

\subsubsection{Object Class Description Examples}\

\begin{table}[h]
\centering
\renewcommand\arraystretch{1.3}
\setlength\tabcolsep{5pt}
\begin{tabular}{lcc}
\hline
\multicolumn{1}{c}{\textbf{Class}}                  & \textbf{Property Description}                                                                                                                                                                                                                                                                                                                             & \textbf{Similarity Description}                                                                                                                                                                                                                                                                                                                     \\ \hline \specialrule{0em}{3pt}{3pt}
\multirow{6}{*}{\textit{Mug}}                       & \begin{tabular}[c]{@{}c@{}}(shape)\textit{``The mug is cylindrical in shape, with a slightly rounded} \\ \textit{base leading  up to straight walls which eventually taper} \\ \textit{slightly towards the rim."}\end{tabular}                                                                                                                                             & \begin{tabular}[c]{@{}c@{}}(similar shape)\textit{``Mugs typically have a cylindrical shape with a} \\ \textit{slightly tapered top and a curved handle; objects with similar} \\ \textit{shapes include bottles and vases."}\end{tabular}                                                                                                                             \\ \specialrule{0em}{3pt}{3pt}
                                           & \begin{tabular}[c]{@{}c@{}}(use)\textit{``A mug is a cylindrical drinking vessel typically used to} \\ \textit{hold  hot beverages such as coffee, tea, hot chocolate,} \\ \textit{or soup. The curved shape of the mug allows liquids} \\ \textit{to be held and consumed while reducing splashes} \\ \textit{and spills."}\end{tabular}                                                      & \begin{tabular}[c]{@{}c@{}}(similar use)\textit{``Mugs and other household objects, such as glasses,} \\ \textit{jars, and other containers, can all be used to hold and contain} \\ \textit{liquids, such as hot or cold drinks. In addition, certain mugs,} \\ \textit{such as those with handles, can also be used to stir and mix} \\ \textit{drinks and other items."}\end{tabular} \\ \specialrule{0em}{3pt}{3pt} \hline \specialrule{0em}{3pt}{3pt}
\multirow{5}{*}{\textit{Spoon}}                     & \begin{tabular}[c]{@{}c@{}}(shape)\textit{``The spoon has a slightly curved bowl shape with a} \\ \textit{long, narrow handle extending diagonally from the bowl in} \\ \textit{the opposite direction to the spoon’s curved edge."}\end{tabular}                                                                                                                            & \begin{tabular}[c]{@{}c@{}}(similar shape)\textit{``A teaspoon and a spoon have a similar shape,} \\ \textit{with both utensils curved at the bowl with a long, thin handle.} \\ \textit{In addition, a soup spoon has a shape similar to a spoon."}\end{tabular}                                                                                                      \\ \specialrule{0em}{3pt}{3pt}
                                           & \begin{tabular}[c]{@{}c@{}}(use)\textit{``A spoon is a unit of measurement for dry and liquid} \\ \textit{ingredients that equals approximately three teaspoons,} \\ \textit{and is used for serving, measuring, and mixing foods."}\end{tabular}                                                                                                                            & \begin{tabular}[c]{@{}c@{}}(similar function)\textit{``A spoon is a utensil commonly used for} \\ \textit{measuring, stirring, and serving food. Household objects} \\ \textit{similar in function include a measuring cup, and soup ladle."}\end{tabular}                                                                                                             \\ \specialrule{0em}{3pt}{3pt} \hline \specialrule{0em}{3pt}{3pt}
\multirow{7}{*}{\textit{Hammer}}                    & \begin{tabular}[c]{@{}c@{}}(shape)\textit{``The hammer has an elongated cylindrical shape with a} \\ \textit{flattened striking face at oneend and a handle at the other end."}\end{tabular}                                                                                                                                                                        & \begin{tabular}[c]{@{}c@{}}(similar shape)\textit{``A hammer has a similar shape as a meat} \\ \textit{tenderizer, an ice pick, and a nail puller."}\end{tabular}                                                                                                                                                                                             \\ \specialrule{0em}{3pt}{3pt}
                                           & \begin{tabular}[c]{@{}c@{}}(use)\textit{``The hammer is a tool used to deliver an impulse force to an} \\ \textit{object. It works by transferring kinetic energy from the handle} \\ \textit{to the striking head of the hammer, allowing the user to drive} \\ \textit{nails, shape and flatten objects, and otherwise manipulate} \\ \textit{the material being worked upon."}\end{tabular} & \begin{tabular}[c]{@{}c@{}}(similar functions)\textit{``A household object that serves a similar} \\ \textit{function to that of a hammer is a kitchen mallet, which is} \\ \textit{used to pound, tenderize, and flatten food. Another} \\ \textit{common household item that serves a similar} \\ \textit{purpose is a rubber mallet.}\end{tabular}                                    \\  \specialrule{0em}{3pt}{3pt}  \hline
\end{tabular}
\caption{Examples of object class descriptions}
\end{table}

\newpage

\subsubsection{Task Description Prompts}
\begin{itemize}
    \setlength{\itemsep}{5pt}
    \item ``\textit{Describe what household objects can be used to [task] in a detailed and scientific response:}"
    \item ``\textit{Describe what household objects support the function of [task] in a detailed and scientific response:}"
    \item ``\textit{Describe what verbs are semantically close to [task] in a detailed and scientific response:}"
    \item ``\textit{Describe what verbs achieve similar effects to ’[task] an object’ in a detailed and scientific response:}" \\
\end{itemize} 

\subsubsection{Task Description Examples}\

\begin{table}[h]
\centering
\renewcommand\arraystretch{1.3}
\setlength\tabcolsep{5pt}
\begin{tabular}{lcc}
\hline
\multicolumn{1}{c}{\textbf{Task}}                   & \textbf{Affordance Description}                                                                                                                                                                                                                                                                                                                           & \textbf{Relevance Description}                                                                                                                                                                                                                                                                                                                      \\ \hline \specialrule{0em}{3pt}{3pt}
\multicolumn{1}{c}{\multirow{5}{*}{\textit{Sweep}}} & \begin{tabular}[c]{@{}c@{}}(use)\textit{``Household objects that can be used to sweep include brooms,} \\ \textit{dustpans and mops.} "\end{tabular}                                                                                                                                                                                                                & \begin{tabular}[c]{@{}c@{}}(closeness)\textit{``Verbs that are semantically close to sweep} \\ \textit{include cleanse, purify, and eradicate."}\end{tabular}                                                                                                                                                                                                 \\  \specialrule{0em}{3pt}{3pt}
\multicolumn{1}{c}{}                       & \begin{tabular}[c]{@{}c@{}}(function)\textit{``Household objects such as a broom, dustpan, and mop help} \\ \textit{to support the sweeping function by providing a tool with which to} \\ \textit{sweep away dust and debris. The broom helps to physically remove} \\ \textit{dirt and dust from the floors and other surfaces."}\end{tabular}                                      & \begin{tabular}[c]{@{}c@{}}(similar effect)\textit{``The action of sweeping an object can be} \\ \textit{described as a type of displacement, in which the object is} \\ \textit{moved across a surface in a steady, sweeping motion. This} \\ \textit{is similar to actions like pushing, dragging, and gliding."}\end{tabular}                                                \\  \specialrule{0em}{3pt}{3pt} \hline  \specialrule{0em}{3pt}{3pt}
\multirow{6}{*}{\textit{Screw}}                     & \begin{tabular}[c]{@{}c@{}}(use)\textit{``Household objects that can be used to screw in objects} \\ \textit{include screwdrivers, managed screwdrivers, and various tools} \\ \textit{with rotating handles, such as adjustable wrenches, pliers,} \\ \textit{and socket wrenches."}\end{tabular}                                                                                    & \begin{tabular}[c]{@{}c@{}}(closeness)\textit{``Verbs that are semantically close to ‘screw’} \\ \textit{might include ‘twist’, ‘fasten’, or ‘tighten’, as these all} \\ \textit{fall within the general semantic domain of attaching or} \\ \textit{fastening objects or components together."}\end{tabular}                                                                   \\  \specialrule{0em}{3pt}{3pt}
                                           & \begin{tabular}[c]{@{}c@{}}(function)\textit{``Household objects such as screwdrivers and} \\ \textit{power drills provide the necessary torque required to} \\ \textit{turn a screw and hold it in place, providing support} \\ \textit{for the function of the screw."}\end{tabular}                                                                                                & \begin{tabular}[c]{@{}c@{}}(similar effect)\textit{``Verbs such as ’fasten’, ’attach’, and ’secure’} \\ \textit{can have similar effects to ’screw an object’, by joining} \\ \textit{two pieces of material together and providing} \\ \textit{a strong connection."}\end{tabular}                                                                                             \\  \specialrule{0em}{3pt}{3pt} \hline  \specialrule{0em}{3pt}{3pt}
\multirow{5}{*}{\textit{Slice}}                     & \begin{tabular}[c]{@{}c@{}}(use)\textit{``Household objects that can be used to slice include knives,} \\ \textit{graters, mandolines, and vegetable peelers. These objects} \\ \textit{typically have blades that are designed to cut through a} \\ \textit{variety of food items."}\end{tabular}                                                                                    & \begin{tabular}[c]{@{}c@{}}(closeness)\textit{``Verbs semantically close to slice include} \\ \textit{chop, divide, section, mince, dice and shred, as} \\ \textit{they all involve cutting an object into smaller pieces,} \\ \textit{either directly or through an intermediary tool."}\end{tabular}                                                                          \\  \specialrule{0em}{3pt}{3pt}
                                           & \begin{tabular}[c]{@{}c@{}}(function)\textit{``Household objects that support the function of} \\ \textit{slicing include knives with sharp edges and fine serrations,} \\ \textit{as well as manual slicers that use an adjustable blade} \\ \textit{to create uniform slices."}\end{tabular}                                                                                        & \begin{tabular}[c]{@{}c@{}}(similar effect)\textit{``Verbs such as cleave, cut, and divide} \\ \textit{can also achieve the same effect as slicing an object,} \\ \textit{by physically splitting the object into two} \\ \textit{or more distinct parts."}\end{tabular}                                                                                                        \\  \specialrule{0em}{3pt}{3pt} \hline
\end{tabular}
\caption{Examples of task descriptions}
\end{table}

\newpage

\subsubsection{Language Instruction Templates}\

\begin{table}[h]
\centering
\renewcommand\arraystretch{2.0}
\begin{tabular}{cc}
\hline
\multicolumn{2}{c}{\textbf{Language Instruction Templates}}                                                                                                                                                                       \\ \hline
\textit{``use the \textless{}obj\textgreater to \textless{}task\textgreater{}"}                                   & \textit{``use the \textless{}obj\textgreater to perform \textless{}tasking\textgreater{}"}                            \\
\textit{``\textless{}task\textgreater things with the \textless{}obj\textgreater{}"}                                & \textit{``use the \textless{}obj\textgreater to \textless{}task\textgreater something"}                               \\
\textit{``executing \textless{}tasking\textgreater with the \textless{}obj\textgreater{}"}                          & \textit{``use the \textless{}obj\textgreater to conduct \textless{}tasking\textgreater{}"}                            \\
\textit{``utilize the \textless{}obj\textgreater to \textless{}task\textgreater{}"}                                 & \textit{``just use the \textless{}obj\textgreater to \textless{}task\textgreater{}"}                                  \\
\textit{``using a \textless{}obj\textgreater to \textless{}task\textgreater{}"}                                     &\textit{``do \textless{}tasking\textgreater with the \textless{}obj\textgreater{}"}                                   \\
\textit{``perform \textless{}tasking\textgreater with the \textless{}obj\textgreater{}"}                            & \textit{``perform \textless{}tasking\textgreater using the \textless{}obj\textgreater{}"}                             \\
\textit{``bring the \textless{}obj\textgreater out to \textless{}task\textgreater{}"}                               & \textit{``to \textless{}task\textgreater{}, get the \textless{}obj\textgreater{}" }                                   \\
\textit{``find the \textless{}obj\textgreater so that you can \textless{}task\textgreater{}"}                       & \textit{``get the \textless{}obj\textgreater and start \textless{}tasking\textgreater{}"},                            \\
\textit{``bring out the \textless{}obj\textgreater to \textless{}task\textgreater{}"}                               & \textit{``perform \textless{}tasking\textgreater with the \textless{}obj\textgreater{}"}                              \\
\textit{``using a \textless{}obj\textgreater to do \textless{}tasking\textgreater{}"}                               & \textit{``make use of the \textless{}obj\textgreater to \textless{}task\textgreater{}"}                               \\
\textit{``grab the \textless{}obj\textgreater to \textless{}task\textgreater{}"}                                    & \textit{``pick up the \textless{}obj\textgreater to \textless{}task\textgreater{}"}                                   \\
\textit{``to \textless{}task\textgreater{}, hold the \textless{}obj\textgreater in your hand"}                      & \textit{``hold the \textless{}obj\textgreater in your hand and \textless{}task\textgreater{}" }                       \\
\textit{``in order to \textless{}task\textgreater{}, grasp the \textless{}obj\textgreater{}"}                       & \textit{``grasp the \textless{}obj\textgreater in order to \textless{}task\textgreater{}"}                            \\
\textit{``if you want to \textless{}task\textgreater{}, hold the \textless{}obj\textgreater{}"}                     & \textit{``grip the \textless{}obj\textgreater to \textless{}task\textgreater{}"}                                      \\
\textit{``\textless{}task\textgreater with the \textless{}obj\textgreater{}"}                                       & \textit{``grasp the \textless{}obj\textgreater in a way that allows for \textless{}tasking\textgreater{}"}            \\
\textit{``holding the \textless{}obj\textgreater in a \textless{}tasking\textgreater{}-friendly manner"}            & \textit{``take a \textless{}tasking\textgreater{}-friendly hold of the \textless{}obj\textgreater{}" }                \\
\textit{``make sure you have a \textless{}tasking\textgreater{}-friendly grip on the \textless{}obj\textgreater{}"} & \textit{``hold the \textless{}obj\textgreater in a \textless{}tasking\textgreater manner"}                            \\
\textit{``grip the \textless{}obj\textgreater in a \textless{}tasking\textgreater{}-friendly manner"}               & \textit{``ensure you have a \textless{}tasking\textgreater{}-friendly grip on the \textless{}obj\textgreater{}"}      \\
\textit{``use the \textless{}obj\textgreater to \textless{}task\textgreater things"}                                & \textit{``performing \textless{}tasking\textgreater with the \textless{}obj\textgreater{}"}                           \\
\textit{``use the \textless{}obj\textgreater to accomplish \textless{}tasking\textgreater{}"}                       & \textit{``\textless{}tasking\textgreater with the \textless{}obj\textgreater{}"}                                      \\
\textit{``do \textless{}tasking\textgreater using the \textless{}obj\textgreater{}"}                                & \textit{``fetch the \textless{}obj\textgreater to \textless{}task\textgreater{}"}                                     \\
\textit{``find the \textless{}obj\textgreater and then \textless{}task\textgreater{}"}                              & \textit{``obtain the \textless{}obj\textgreater for \textless{}tasking\textgreater{}"}                                \\
\textit{``use the \textless{}obj\textgreater to conduct \textless{}tasking\textgreater{}"}                          & \textit{``grasp the \textless{}obj\textgreater to \textless{}task\textgreater{}"}                                     \\
\textit{``taking hold of the \textless{}obj\textgreater{}, \textless{}task\textgreater{}"}                          & \textit{``to \textless{}task\textgreater{}, grasp the \textless{}obj\textgreater{}"}                                  \\
\textit{``to \textless{}task\textgreater{}, take hold of the \textless{}obj\textgreater{}"}                         & \textit{``get the \textless{}obj\textgreater to \textless{}task\textgreater{}"}                                       \\
\textit{``grasp the \textless{}obj\textgreater in a way that allows you to \textless{}task\textgreater{}"}          & \textit{``ensure you grasp the \textless{}obj\textgreater in a way that allows for \textless{}tasking\textgreater{}"} \\
\textit{``hold the \textless{}obj\textgreater in a \textless{}tasking\textgreater position"}                        &                                                                                                             \\ \hline
\end{tabular}
\caption{Language instruction templates}
\end{table}

\newpage

\subsubsection{Dataset Examples} \

\begin{figure}[h]
  \centering
  \begin{tikzpicture}[inner sep = 0pt, outer sep = 0pt]
    \node[anchor=south west] (fnC) at (0in,0in)
      {\includegraphics[height=6.8in,clip=true,trim=0in 0in 0in 0in]{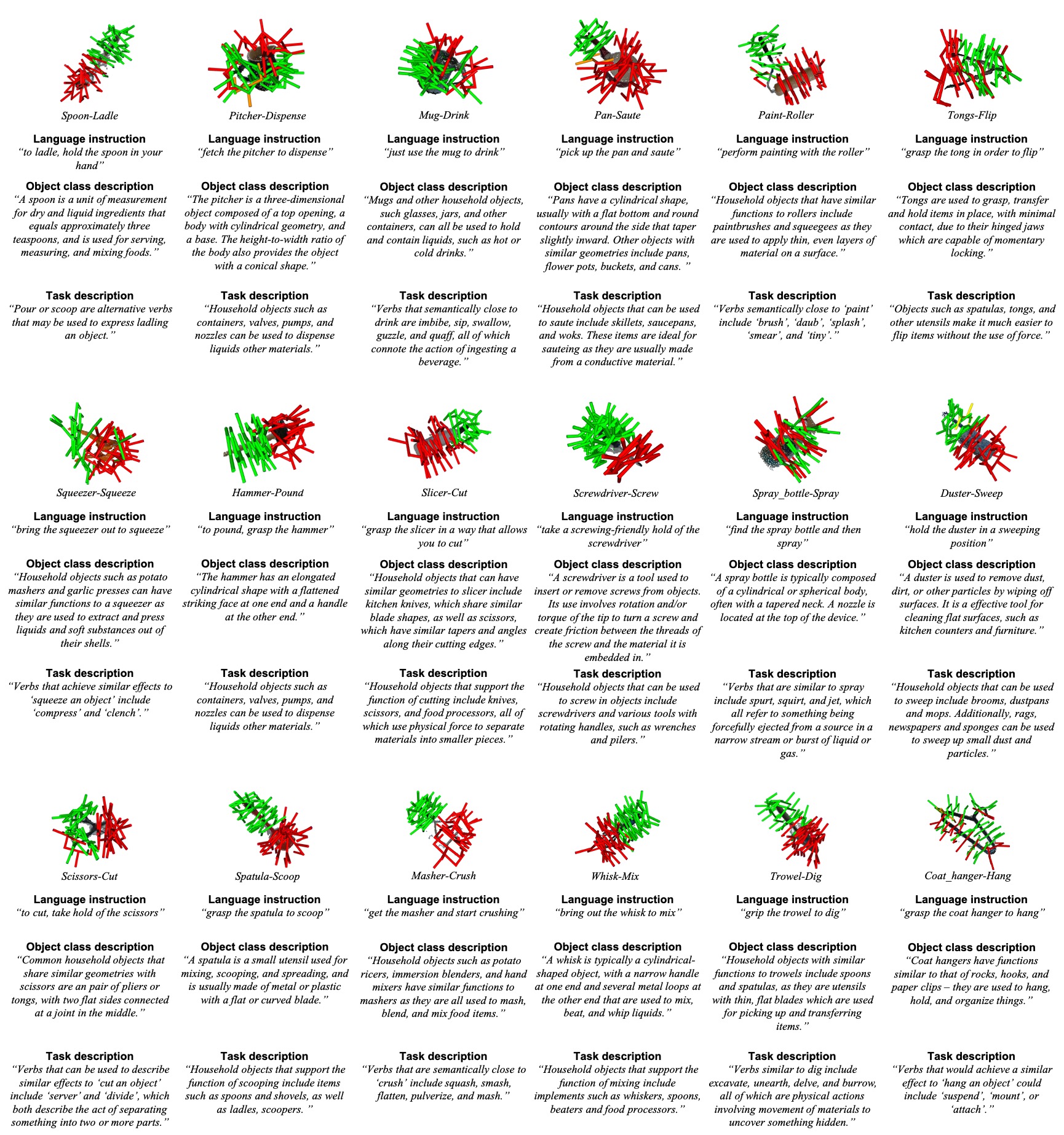}};
  \end{tikzpicture}
  \caption{Dataset examples, each of which includes 6 DoF task-oriented grasp poses, a language instruction, an object class description paragraph, and a task description paragraph. Here, we only show part of the paragraphs. All the grasp poses are colored by their task compatibility scores (green is higher).} 
\end{figure}

\newpage

\subsection{Additional Experimental Setup}
This section aims to provide further information regarding the experimental setup employed for both perception and real-robot experiments. \\

\subsubsection{Training Hyper-Parameters} \

\begin{table}[h]
\renewcommand\arraystretch{1.5}
\centering
\begin{tabular}{lc}
\hline
\multicolumn{2}{l}{\textbf{Basic Setting}}                      \\ \hline
Batch Size           & 32                              \\
\# of Points         & 4096                            \\
\# of Epochs         & 50                              \\ \hline
\multicolumn{2}{l}{\textbf{Optimization Setting}}               \\ \hline
Optimizer            & Adam                            \\
Learning Rate                  & 0.0001                          \\
Learning Rate Decay             & 0.7                             \\
Decay Step           & 2e4                             \\
Weight Decay         & 0.0001                          \\
Learning Rate Clip              & 1e-5                            \\ \hline
\multicolumn{2}{l}{\textbf{PointNet++ Setting}}                 \\ \hline
\# of SA Layers      & 3                               \\
\# of Sampled Points & 512, 128, 1                     \\
Embedding Sizes      & 320, 640, 1024                  \\ \hline
\multicolumn{2}{l}{\textbf{Data Preprocessing}}                 \\ \hline
Scaling              & True                            \\
Mean Centering       & True                            \\
Random Rotation      & True                            \\
Random Jitter        & True                            \\
Random Dropout       & True                            \\ \hline
\multicolumn{2}{l}{\textbf{Hardware Resource}}                  \\ \hline
CPU                  & 12th Gen Intel® Core™ i9-12900K \\
\# of  CPU Cores     & 24                              \\
GPU                  & Nvidia RTX 3090                 \\ \hline
\multicolumn{2}{l}{\textbf{LLM Setting}}                  \\ \hline
Model              & OpenAI GPT-3     \\
Engine             & \textit{text-davinci-003} \\
Prompt Type             & text             \\
Temperature        & 1.0              \\
Max Tokens        & 256              \\
Top P             & 1.0              \\
Frequency Penalty & 0.0              \\
Presence Penalty  & 0.0              \\ \hline
\end{tabular}
\caption{Training hyper-parameter setting}
\end{table}

\newpage

\subsubsection{Real-Robot Experiment} \ 

\begin{figure}[th]
  \centering
  \vspace*{-0.2in}
  \begin{tikzpicture}[inner sep = 0pt, outer sep = 0pt]
    \node[anchor=south west] (fnC) at (0in,0in)
      {\includegraphics[height=1.5in,clip=true,trim=0in 0in 0in 0in]{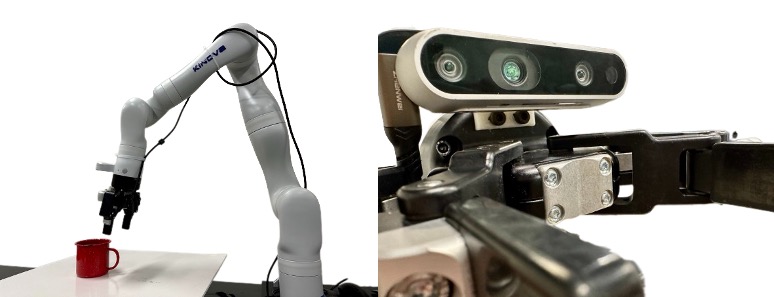}};
  \end{tikzpicture}
  \caption{Real-robot experimental setup: a Kinova Gen3 robotic arm with a Robotiq parallel jaw gripper (left) and an Intel RealSense D435 RGB camera (right) with eye-in-hand calibration.}
\end{figure}

\begin{figure}[h]
  \centering
  \vspace*{-0.1in}
  \begin{tikzpicture}[inner sep = 0pt, outer sep = 0pt]
    \node[anchor=south west] (fnC) at (0in,0in)
      {\includegraphics[height=2.2in,clip=true,trim=0in 0in 0in 0in]{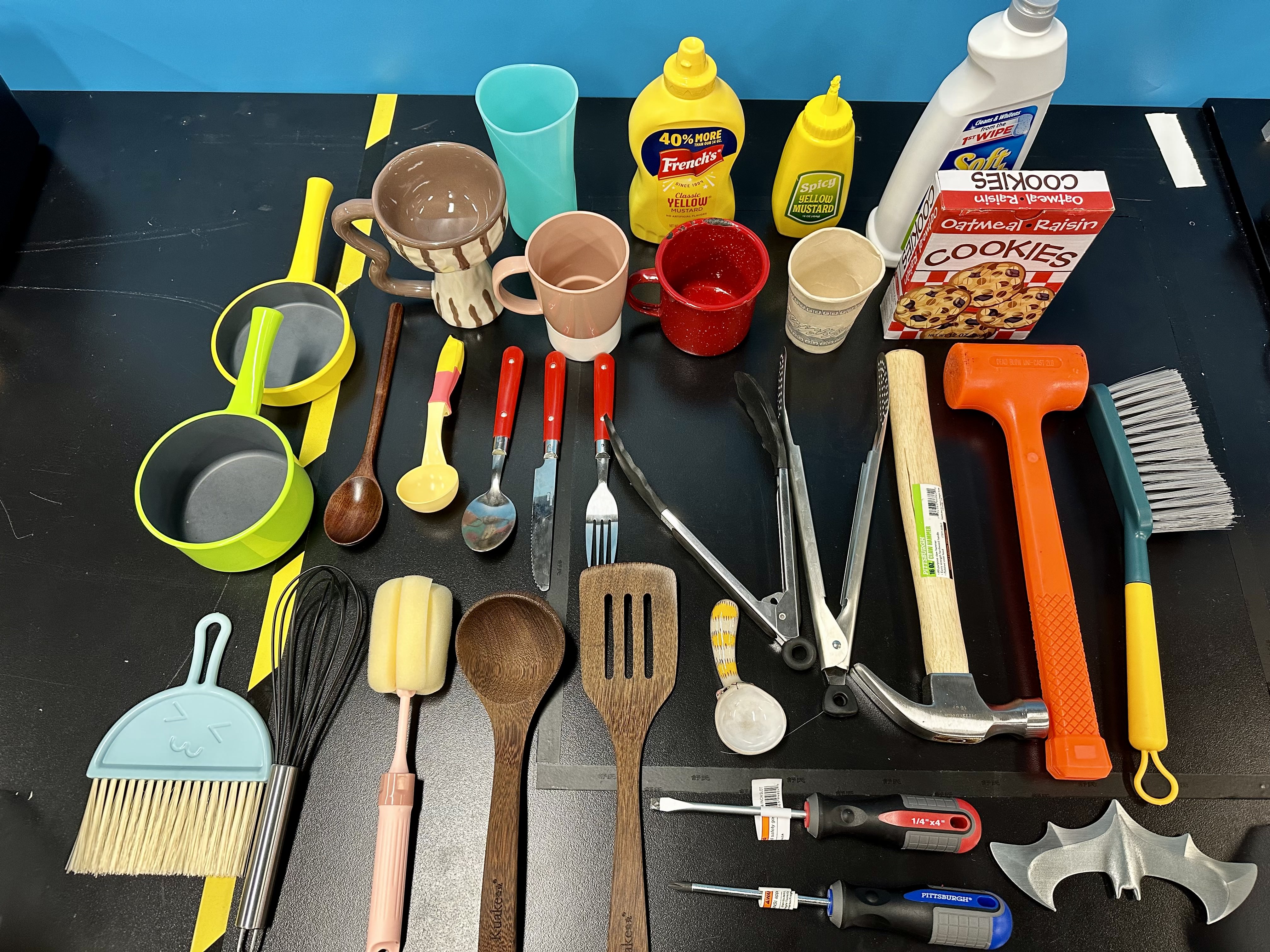}};
  \end{tikzpicture}
  \caption{Test objects collected from our laboratory and YCB dataset.}
\end{figure}

\begin{figure}[h]
  \centering
  \vspace*{-0.2in}
  \begin{tikzpicture}[inner sep = 0pt, outer sep = 0pt]
    \node[anchor=south west] (fnC) at (0in,0in)
      {\includegraphics[height=2.5in,clip=true,trim=0in 0in 0in 0in]{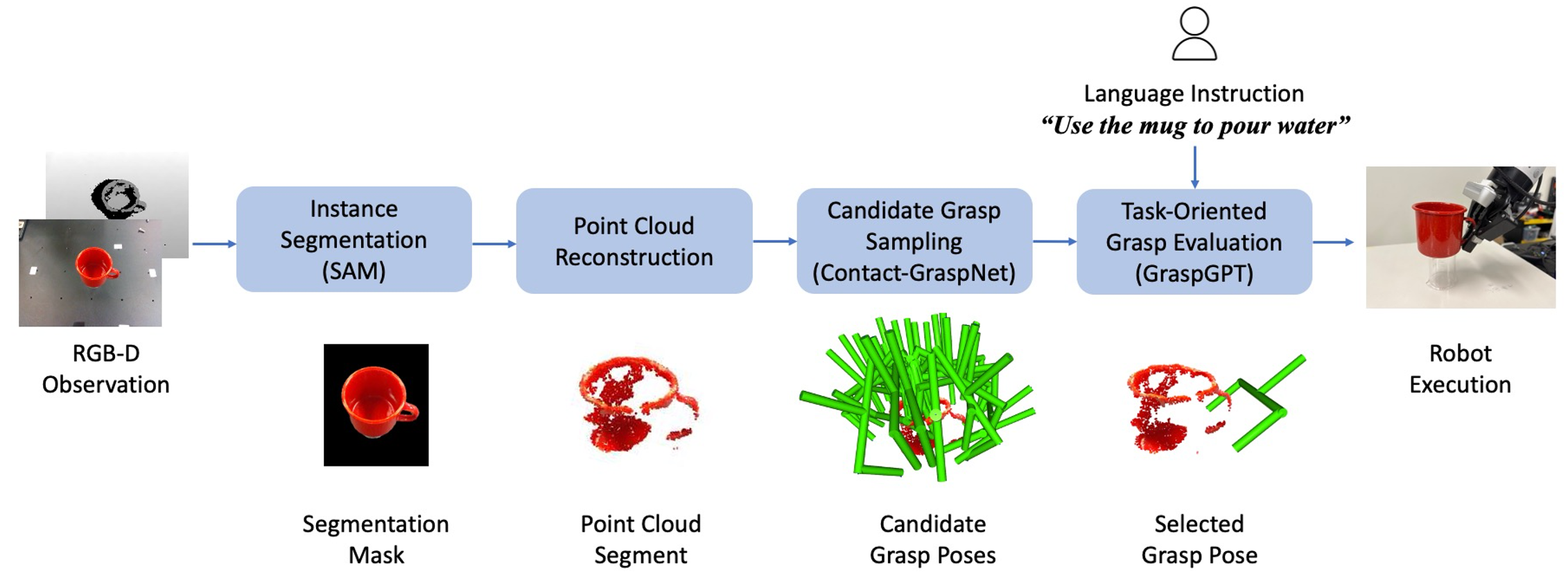}};
  \end{tikzpicture}
  \caption{Pipeline of real-robot experiment with intermediate results.}
\end{figure}

\newpage





\subsection{Discussion}
In this section, we discuss the limitations of GraspGPT. Potential solutions are also provided as part of our future work.
\begin{itemize}
    \setlength{\itemsep}{5pt}
    \item \textbf{LLM Knowledge Filtering and Selection}  

    As previously mentioned, we do not process the language data returned by an LLM, which can result in language descriptions containing imprecise or false commonsense knowledge. We identify two typical errors in the generated language descriptions: (1) Part-of-speech error. Since certain words have multiple uses as nouns and verbs, LLM occasionally returns object class knowledge even when prompted with task description prompts, or vice versa. For instance, when we prompt the LLM for the task description of verb \textit{``ladle"} (e.g., \textit{``Describe what household objects support the function of `ladle' in a detailed and scientific response:"}), the LLM might respond with \textit{``A ladle is a utensil that is typically long-handled, with a deep dish or scoop at the end. It is usually made of metal or plastic and is used to serve or measure hot liquids such as soup, sauce, or gravy."} (2) Mismatching descriptions. Due to significant intra-class variances, the LLM-generated object class descriptions may not precisely match the properties of actual object instances. For example, the object class description of \textit{``frying pan"} could be:  \textit{``The geometry of a frying pan is generally cylindrical, with sloping sides and a flat base to evenly disperse heat while cooking."} However, the actual object instance might be square-shaped. To address the first error, GraspGPT should be able to inspect the semantic meaning of the generated language description and verify if it meets the prompt's intention. To tackle the second error, a potential solution would involve incorporating a multi-modal model, such as \href{https://arxiv.org/abs/2301.12597}{BLIP-2}, which can generate language descriptions based on the visual content provided. This integration would require GraspGPT to process additional visual inputs, such as RGB images.
        
    \item \textbf{Task-Oriented Pick and Place} 
    
    Our current work focuses on addressing the challenge of task-oriented grasping/picking. However, to achieve successful tool manipulation, the robot must also anticipate the subsequent motion of the tool and effectively interact with the target object. For instance, when inserting a nail into a slot, the robot needs to perform the following steps: (1) securely grasp the hammer by its handle, (2) guide the hammer towards the nail, assuming the nail is initially positioned halfway inside the slot, and (3) forcefully pound the nail into the slot. While GraspGPT has progressed in addressing the initial task-oriented grasping step, the subsequent steps are currently simplified with pre-defined motion primitives. Recent works, such as \href{https://arxiv.org/abs/2112.05124}{Neural Descriptor Fields}, approach tool manipulation as a pick-and-place task. This involves predicting the grasp point on the tool object (i.e., task-oriented picking) and determining the effect point on the target object (i.e., task-oriented placing). Since we currently use rule-based heuristics to determine the effect point, the robot cannot model the relative pose between the tool object and the target object. Ideally, the robot should adjust the effect point depending on the grasp point. A failure case can be found in the supplementary video. We plan to expand the capabilities of GraspGPT from task-oriented grasping to task-oriented pick and place, leveraging both grasping and placement knowledge from an LLM.
        
    \item \textbf{Single-Stage Architecture} 
    
    GraspGPT currently follows a two-stage, sample-and-evaluate approach, similar to previous works. While this design choice simplifies the complexity of constructing GraspGPT, it introduces a reliance on a pre-trained task-agnostic grasp sampler. As the grasp sampler solely considers geometry information without incorporating semantic priors about the tool to be grasped, it uniformly samples over the given point cloud. However, during robot interaction, only specific functional/affordance regions of a tool are engaged, while non-functional regions remain untouched. This uniform sampling approach makes GraspGPT inefficient for real-time inference. Moreover, GraspGPT assumes all the candidate grasp poses generated by the grasp sampler are stable. However, the sampler may output marginal/unstable grasp poses, which are susceptible to perturbances such as unpredicted contact or calibration error. An example can be found in the supplementary video. Future research aims to integrate task-agnostic sampling and task-oriented evaluation within an end-to-end architecture, enabling the direct prediction of task-oriented grasp poses from given point clouds. This approach is anticipated to consider both stability and task compatibility simultaneously. 
    
    \item \textbf{Simultaneous Affordance Learning} 
    
     A closely related task to task-oriented grasping is affordance recognition, where the robot identifies specific regions on an object for various types of interactions. Previous studies, such as the \href{https://ieeexplore.ieee.org/abstract/document/9364360}{Affordance Keypoint Detection Network} (\textit{AffKP}), have demonstrated the benefits of joint learning of affordance segmentation for task-oriented grasping and manipulation. In future work, we aim to incorporate affordance learning into the GraspGPT framework. We anticipate that simultaneously learning these two objectives would mutually enhance their performance. On the one hand, affordance learning would assist the robot in identifying the relevant regions to grasp for a given task. On the other hand, the supervision provided by task-oriented grasping could serve as weak supervision for affordance recognition. Equipping the robot with affordance recognition capability also opens up possibilities for other tasks, such as task-driven object retrieval/selection and semantic scene understanding.

\end{itemize}

\end{appendices}

\end{document}